\documentclass[sigconf]{acmart}

\AtBeginDocument{%
  \providecommand\BibTeX{{%
    \normalfont B\kern-0.5em{\scshape i\kern-0.25em b}\kern-0.8em\TeX}}}

\setcopyright{acmcopyright}
\copyrightyear{2022}
\acmYear{2022}
\acmDOI{10.1145/3503161.3547869}

\acmConference[MM'2022]{ACM MULTIMEDIA CONFERENCE 2022}{October 10--14, 2022}{Lisbon, Portugal}
\acmPrice{15.00}
\acmISBN{978-1-4503-9203-7/22/10}

\acmSubmissionID{496}

\usepackage{multirow}
\usepackage{enumitem}
\usepackage{bbding}
\usepackage[capitalize]{cleveref}
\crefname{section}{Sec.}{Secs.}
\Crefname{section}{Section}{Sections}
\Crefname{table}{Table}{Tables}
\crefname{table}{Tab.}{Tabs.}

\begin{document}

\title{MM-Pyramid: Multimodal Pyramid Attentional Network for Audio-Visual Event Localization and Video Parsing}

\author{Jiashuo Yu\textsuperscript{1}, Ying Cheng\textsuperscript{2}, Rui-Wei Zhao\textsuperscript{2},  Rui Feng\textsuperscript{1,2,3*}\authornotemark[1], Yuejie Zhang\textsuperscript{1,3*}\authornotemark[1]}

\affiliation{
\textsuperscript{1}School of Computer Science, Shanghai Key Laboratory of Intelligent Information Processing, Fudan University, China
\\\textsuperscript{2}Academy for Engineering and Technology, Fudan University, China
\\\textsuperscript{3}Shanghai Collaborative Innovation Center of Intelligent Visual Computing, China
\country{}}
\email{{jsyu19,chengy18,rwzhao,fengrui,yjzhang}@fudan.edu.cn}

\begin{abstract}
Recognizing and localizing events in videos is a fundamental task for video understanding. Since events may occur in auditory and visual modalities, multimodal detailed perception is essential for complete scene comprehension. Most previous works attempted to analyze videos from a holistic perspective. However, they do not consider semantic information at multiple scales, which makes the model difficult to localize events in different lengths. In this paper, we present a Multimodal Pyramid Attentional Network (\textbf{MM-Pyramid}) for event localization. Specifically, we first propose the attentive feature pyramid module. This module captures temporal pyramid features via several stacking pyramid units, each of them is composed of a fixed-size attention block and dilated convolution block. We also design an adaptive semantic fusion module, which leverages a unit-level attention block and a selective fusion block to integrate pyramid features interactively. Extensive experiments on audio-visual event localization and weakly-supervised audio-visual video parsing tasks verify the effectiveness of our approach.
\end{abstract}

%\ccsdesc[500]{Computing methodologies~Activity recognition and understanding}

\maketitle

\renewcommand{\thefootnote}{\fnsymbol{footnote}}
\footnotetext[1]{indicates corresponding authors.}


\section{Introduction}

\begin{figure}[t]
\centering
\includegraphics[width=0.47\textwidth]{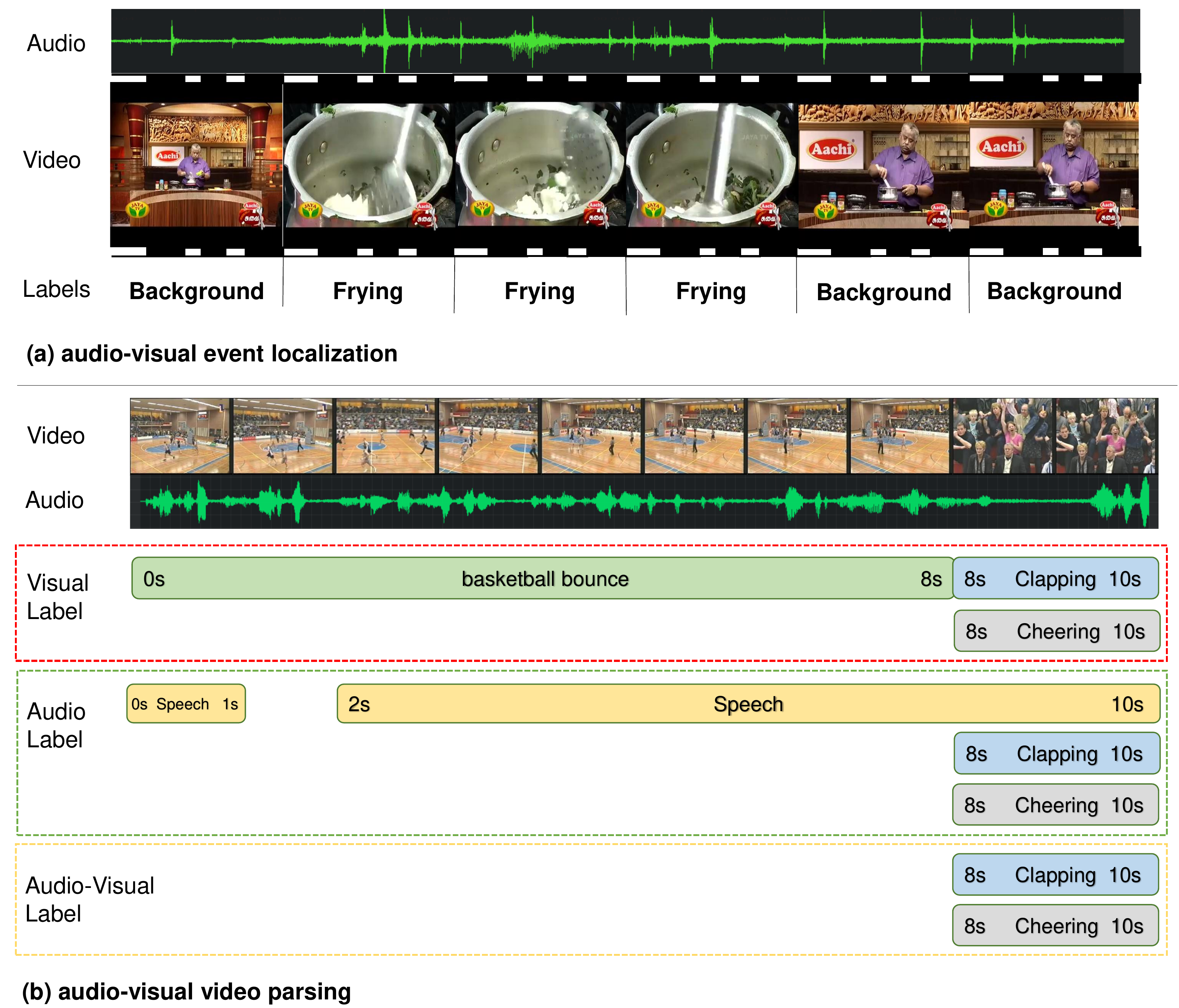}
\vspace{-2mm}
\caption{Audio-visual event localization aims to temporally localize a given audio-visual event, while audio-visual video parsing task requires to classify and localize all uni-modal and multimodal events in different lengths.}
\label{figure1}
\vspace{-4mm}
\end{figure}

Video scene understanding in computer vision is fundamental for many real-world applications and it simulates the information perception process of human brain. According to the researches in cognitive neuroscience~\cite{bulkin2006seeing,goller2009seeing}, human perceives information from multiple modalities to obtain the overall comprehension. Similar to human brains, auditory and visual data can provide complementary cues from different perspectives for machine video understanding.  

In recent years, some works~\cite{korbar2018cooperative,owens2018audio,arandjelovic2017look, cheng2020look, gan2019self, alayrac2020self, ma2021active, morgado2020learning, alwassel2020self} focus on the synergistic effects between auditory and visual modalities and acquire the joint multimodal representation, and some other works~\cite{Zhao_2018_ECCV, zhao2019sound, hu2020discriminative, Afouras20b} investigate on localizing sounding objects via self-supervised methods. However, the analysis of audio-visual events in videos, which is a crucial part of the video scene perception, is also in need of investigation. To this end, some tasks and corresponding methods are proposed to explore the impact of audio-visual cues on events. Specifically, Tian et al.~\cite{tian2018audio} propose the audio-visual event localization task, which aims to classify and temporally localize an audio-visual event in a video clip. As shown in~\cref{figure1}(a), the audio-visual event \textit{frying} cannot be seen in the first and last two segments, thereby labeled as \textit{background}. In the other seconds, the food frying can both be heard and seen, thus we label these as~\textit{frying}. To make this task more generalizable, Tian et al.~\cite{tian2020avvp} expand the task of localizing one event to multiple events scenarios and introduce the audio-visual video parsing task, which is illustrated in~\cref{figure1}(b), given a video that includes several audible, visible, and audi-visible events, the audio-visual video parsing task aims to predict all event categories, distinguish the modalities perceiving each event, and localize their temporal boundaries. Since the process of labeling all event boundaries is cumbersome, this task is conducted in a weakly-supervised manner, which makes it more generalizable to real-world applications yet more challenging.

Some researchers tackle these problems by capturing contexts from a holistic perspective. For audio-visual event localization, prior works~\cite{tian2018audio, wu2019dual, lin2019dual, xuan2020cross, ramaswamy2020makes, zhou2021positive, CMRAN2020Xu} explore the relationship between auditory and visual sequences via different kinds of attention mechanisms. For audio-visual video parsing, Tian et al.~\cite{tian2020avvp} propose a hybrid attention network to capture temporal context, which tends to focus more on the holistic content and is capable to detect the major event throughout the video. However, these methods are limited by some cases including when the lengths of target events are short, or videos include several events that have miscellaneous lengths. Since they focus more on the coarse-grained holistic content, detailed information is inclined to be neglected, which makes it difficult to localize short-term events. Despite several methods~\cite{wang2017spatiotemporal, zhang2018dynamic, yang2020temporal} proposed to capture temporal pyramid features, they can only tackle uni-modal scenarios and lack multimodal interactions. Therefore, the necessity of exploring features both in different granularities and modalities emerges, which helps to localize multimodal events in different temporal sizes accurately and further leads to a comprehensive video understanding.

\begin{figure*}[t]
\centering
\includegraphics[width=\textwidth]{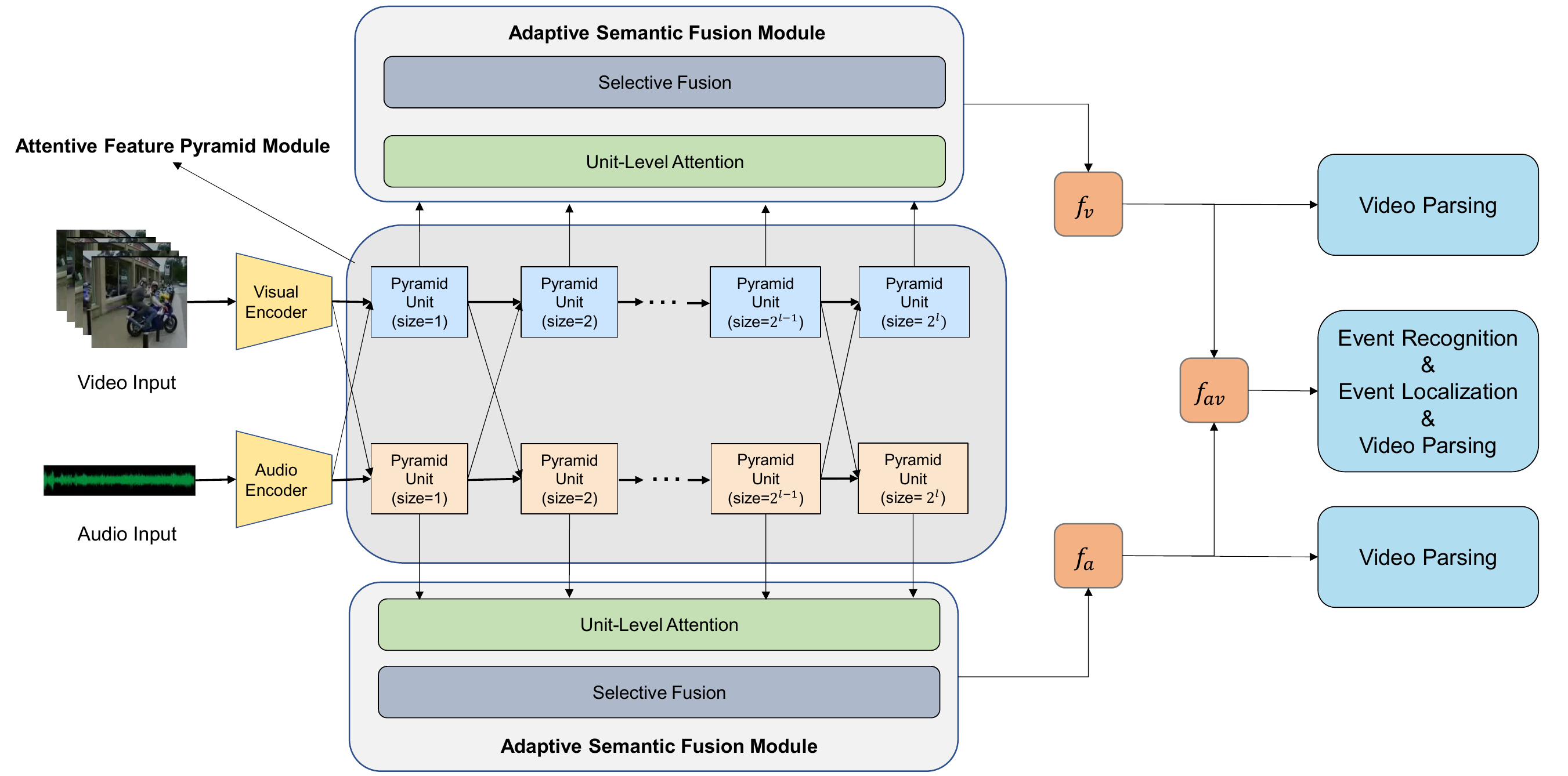}
\caption{An overview of our proposed Multimodal Pyramid Attentional Network (MM-Pyramid). Our proposed framework consists of two parts: the attentive feature pyramid module and the adaptive semantic fusion module. Take the features extracted from pretrained networks as input, the attentive feature pyramid module captures multimodal pyramid features by multiple pyramid units in different scales. The adaptive semantic fusion module integrates pyramid features via the unit-level attention and the selective fusion operation.}
\label{figure2}
\end{figure*}

In this paper, we introduce a novel Multimodal Pyramid Attentional Network \textbf{(MM-Pyramid)}. To be specific, we first propose a novel attentive feature pyramid module composed of multiple pyramid units to acquire multi-level audio-visual features. In each pyramid unit, a fixed-size multi-scale attention block captures intra- and inter-modality interactions, together with a dilated convolution block to integrate segment-wise features and derive semantic information. To fuse pyramid units, we also design an adaptive semantic fusion module. This module explores the correlations among multi-level features and integrates pyramid units in a selective fusion way. By this means, the model can obtain more targeted representation, thereby resulting in better audio-visual event localization and video parsing performance. In summary, our contributions are as follows: 

\begin{itemize}[leftmargin=*]
\item We propose to exploit audio-visual pyramid features to learn multi-scale semantic information and localize events in different lengths, which is beneficial for a comprehensive video scene understanding.
\item We develop a novel Multimodal Pyramid Attentional Network, which consists of an attentive feature pyramid module and an adaptive semantic fusion module to capture and integrate multi-level features, respectively.
\item We conduct extensive experiments on two audio-visual tasks: audio-visual event localization on the AVE~\cite{tian2018audio} dataset and weakly-supervised audio-visual video parsing on the LLP~\cite{tian2020avvp} dataset to verify the effectiveness of our proposed framework.
\end{itemize}

\section{Related Works}

\subsection{Audio-visual representation learning.}
Audio-visual representation learning aims to acquire the informative multimodal representation by exploiting the correlations between auditory and visual modalities. Some works~\cite{arandjelovic2017look, korbar2018cooperative, owens2018audio, cheng2020look, gan2019self, alayrac2020self, ma2021active, morgado2020learning} try to obtain the joint audio-visual representation by learning the correspondence of audio and visual streams in a self-supervised manner. Others~\cite{alwassel2020self, hu2019deep} leverage unsupervised clustering as the supervision to explore the cross-modal correlation. Besides, some other works~\cite{Zhao_2018_ECCV, zhao2019sound, gan2019self, gan2020music, Afouras20b} explore the relationship between the sound and dynamic motions of objects and enhance the capability of object localization. In this paper, we try to leverage the correlation between audio and visual content to enhance the performance of downstream applications.

\subsection{Audio-visual event localization and video parsing.}
Audio-visual event localization~\cite{tian2018audio} utilizes the synergy and relevance between auditory and visual streams to temporally localize events in the given video. Most prior works~\cite{wu2019dual, lin2019dual, tian2018audio, ramaswamy2020makes, xuan2020cross, CMRAN2020Xu} leverage the attention-based architecture to capture inter- and intra-modality interactions for holistic video understanding. Yu et al.~\cite{yu2021mpn} explores the differences of video-level classification and segment-level localization, and propose a multimodal parallel network to decrease the conflicts between global and local features. More recently, Zhou et al.~\cite{zhou2021positive} proposes a positive sample propagation strategy to utilize positive audio-visual pairs, thereby learning discriminative features for the classifier. 

To expand the event localization task to multi-event scenarios, Tian et al.~\cite{tian2020avvp} propose a more generalizable and challenging task named audio-visual video parsing, which aims to classify and locate all audible, visible, and audi-visible events inside a video in a weakly-supervised manner. They also propose a hybrid attention network to capture multimodal contexts and a multimodal multiple instance learning method for the weakly supervised setting. Wu et al.~\cite{wu2021exploring} propose to obtain accurate modality-aware event supervision by swapping audio and visual tracks with other unrelated videos to address the modality uncertainty issue. In this paper, we propose to explore multi-scale audio-visual features for localizing events in multiple lengths, which is neglected by previous methods.

\section{Task formulation.} 
\label{sec3.1}

\noindent\textbf{Audio-visual event localization} aims to classify and localize an audio-visual event in a given video. The task can be tackled in the fully-supervised and weakly-supervised manners. For the fully-supervised setting, the event label for the $t_{th}$ video segment is given as $y_p=\{y_t^p|y_t^p\in\{0,1\},p=1,...,C,\sum_{p=1}^C y_t^p = 1\}$, where $C$ is the total number of audio-visual events plus one background category, while in the weakly-supervised setting, only video-level event categories are given during training yet temporal boundaries are still required during inference.

\noindent\textbf{Weakly-supervised audio-visual video parsing} aims to predict all event categories, distinguish the modalities perceiving each event, and localize their temporal boundaries. Given a video sequence $\{V_t, A_t\}_{t=1}^N$ with $N$ non-overlapping temporal segments, the event labels are given as $y_t=\{(y_t^v, y_t^a, y_t^{av})|[y_t^v]^m, [y_t^a]^m, [y_t^{av}]^m\in \{0,1\}, m = 1, ..., C\}$, where $C$ is the total number of event categories. An event is labeled as audi-visible only when it is both audible and visible, thus the audi-visible label can be computed as $y_t^{av} = y_t^v * y_t^a$. This task is conducted in a weakly supervised manner. We only have all event categories that appeared in the given video for training, but need to predict which segments contain those events and which modalities perceive them during inference. 

\section{Methodology}

In this section, we introduce our Multimodal Pyramid Attentional Network, which is shown in~\cref{figure2}. We first propose the attentive feature pyramid module to obtain temporal pyramid features, which is introduced in \cref{subsec: attentive}. Then we propose an adaptive semantic fusion module for an interactive pyramid feature fusion in \cref{subsec: adaptive}, respectively.

\subsection{Attentive Feature Pyramid Module}
\label{subsec: attentive}
\begin{figure}[t]
\centering
\includegraphics[width=0.47\textwidth]{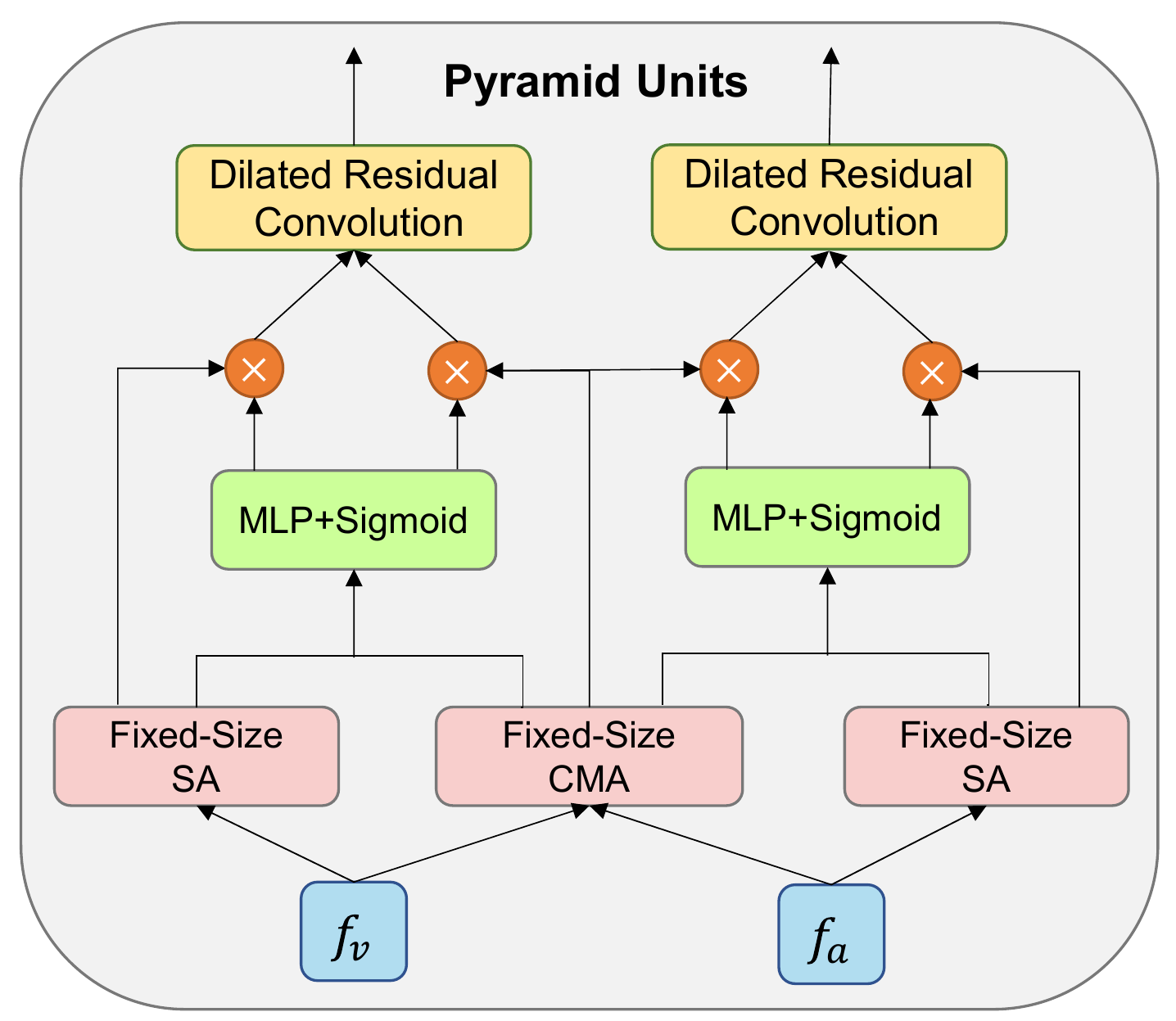}
\caption{Detailed structure of two linked audio and visual pyramid units in the same level. $\times$ is the channel-wise multiplication. SA and CMA denote self-attention and cross-modal attention.}
\label{figure3}
\end{figure}

The attentive feature pyramid module is composed of a few stacked units in different scales. Pyramid units in different modalities are connected interactively. The detailed structure of two linked audio and visual pyramid units in the same size are shown in~\cref{figure3}. In each unit, we first propose the fixed-size attention mechanism to introduce intra- and inter-modality interactions, then perform feature integration via a dilated residual convolution block. The size of each unit is different and the outputs of all units are preserved as pyramid-like multimodal features. 

\noindent\textbf{Attentive feature interaction.} Self-attention (SA) and cross-modal attention (CMA) are used to provide temporal feature interactions. We reform the encoder part of Transformer~\cite{vaswani2017attention}. Specifically, the attention scores between different video snippets is computed by the scaled dot-product attention $att(q, k, v) = softmax(\frac{qk^T}{\sqrt{d_m}})v$, where $q, k, v$ denotes the query, key, and value vectors, $d_m$ is the dimension of query vectors, $T$ denotes the matrix transpose operation. The self-attention block learns uni-modal temporal relationships via $sa(f) = att(fW_{q}, fW_{k}, fW_{v})$, where $W_{q}, W_{k}, W_{v}$ are learnable parameters, $f$ is the input feature. For the cross-modal attention block, we assign features in current modality as the query vectors, while the key and value vectors are from features of the other modality. The formulations can be defined as $cma(f_v, f_a) = att(f_vW_{q}, f_aW_{k}, f_aW_{v})$ and $cma(f_a, f_v) = att(f_aW_{q}, f_vW_{k}, f_vW_{v})$, where $f_a$ is the audio feature, $f_v$ is the video feature, $W_{q}, W_{k}$, and $W_{v}$ are learnable parameters. The parameter matrices in the cross-modal attention block are shared. This parameter-efficient setting can project audio and visual features into the same subspaces, which facilitates further interactions of uni-modal and multimodal features. Then the features are processed by a feed-forward layer. We adopt the layer normalization~\cite{ba2016layer} for regularization, and the residual connections~\cite{he2016deep} for the identity mapping to avoid overfitting.

\begin{figure}[t]
\centering
\includegraphics[width=0.47\textwidth]{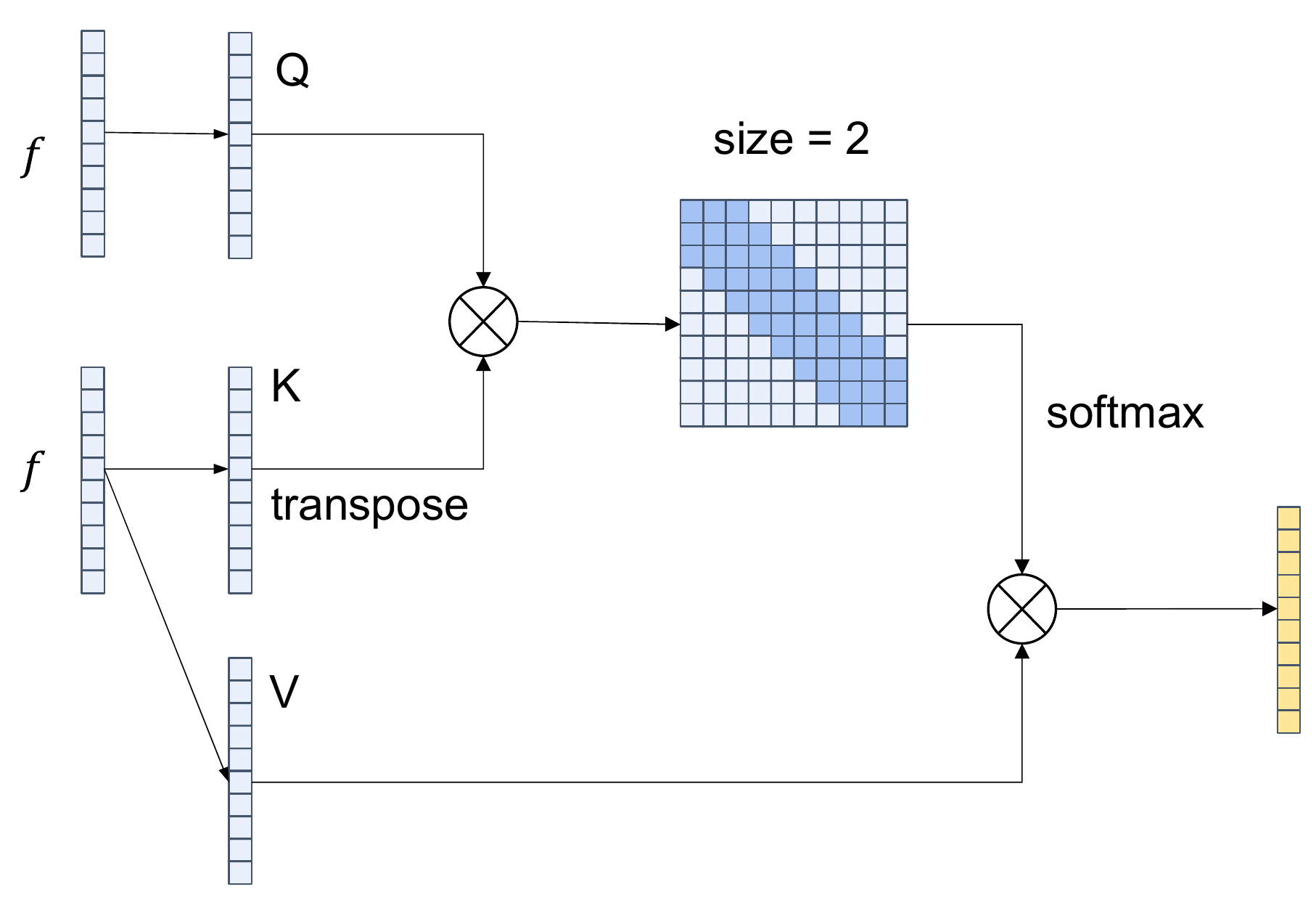}
\caption{Detailed structure of the fixed-size attention mechanism (the size of the interaction window is 2).}
\label{figure4}
\end{figure}

Since we intend to obtain pyramid features for localizing different lengths of events, the temporal interacting size of each unit ought to be diverse. To this end, we set an interaction window to constrain the interacting size of the self-attention and cross-modal attention layer. Concretely, we propose a fixed-size attention block as shown in~\cref{figure4}, which restricts the interaction windows by adding masks to areas that should not be involved. In this way, the fixed-size attention can be computed as follows,
\begin{equation}
\small
    sa(f, d) = att(fW_{q}, S_t(f, d)W_{k}, S_t(f, d)W_{v}),
\end{equation}
\begin{equation}
\small
    cma(f_v, f_a, d) = att(f_vW_{q}, S_t(f_a, d)W_{k}, S_t(f_a, d)W_{v}),
\end{equation}
\begin{equation}
\small
    cma(f_a, f_v, d) = att(f_aW_{q}, S_t(f_v, d)W_{k}, S_t(f_v, d)W_{v}),
\end{equation}
\begin{equation}
\small
    S_t(x, d) = [x_{t-d},...,x_{t+d}],
\end{equation}
where $S(t)$ indicates creating interaction windows for the $t^{th}$ segment, $d$ denotes the size of the interaction window.

Different from some prior works~\cite{cheng2020look, tian2020avvp} where self-attention and cross-modal attention blocks are connected in serial, we adopt a parallel arrangement. The inputs of the two kinds of attention blocks are both the output of the previous pyramid unit. We then leverage the channel-wise attention to interconnect and integrate uni-modal and multimodal features. To be specific, the output of self-attention and cross-modal attention blocks are firstly concatenated along the channel dimension. Then channel-wise attention scores are computed to reﬁne raw features via a linear layer followed by a sigmoid function. The final fused features are calculated by the summation of the refined uni-modal and multimodal features. The fusion process of visual modality is formulated as below,
\begin{equation}
\small
    F^v_{fused} = \sigma(W_{sa}F^v_c+b_1)F^v_{sa} + \sigma(W_{cma}F^v_c+b_2)F^v_{cma},
\end{equation}
\begin{equation}
\small
    F^v_c = Concat(F^v_{sa}, F^v_{cma}),
\end{equation}
where $W_{sa}, W_{cma}, b_1, b_2$ are learnable weights. The formulation of auditory modality fusion is highly similar, thus we omit it for concise writing.

\noindent\textbf{Dilated temporal convolution.} Directly utilizing outputs of the fixed-size attention will lead to two problems: First, though interactions among temporal segments have been performed sufficiently, features still need to be amalgamated in temporal dimension to perceive semantic information. Second, since positional encoding is not performed in the attention blocks, temporal order of the sequence has not been modeled, which is important for event understanding. Therefore, a temporal convolution block is used to inject positional information and derive semantic representation.

Temporal convolutional network has been widely applied in speech synthesis~\cite{oord2016wavenet} and action segmentation~\cite{lea2017temporal,farha2019ms,li2020ms}. They can provide multi-grained information via multiple dilated convolution layers. The dilation size of each layer is increasing exponentially, which expands the receptive fields at each layer, thus the network can focus on information in distinct temporal lengths. Following~\cite{li2020ms}, we adopt the dilated residual block for our temporal convolution block. Each dilated residual block contains a $3\times3$ dilated convolution, a ReLU~\cite{glorot2011deep} activation, a $1\times1$ convolution, and the residual connection. Moreover, instead of causal convolution adopted in some temporal forecasting tasks, we use acausal convolution with kernel size 3 since it can take more contextual information of the current segment into consideration. The operations in each dilated residual block can be described as follows,

\begin{equation}
\small
\hat{F_t^l} = ReLU(W^1F_t^{l} + W^2F_{t-d}^{l} + W^3F_{t+d}^{l} + b_3),
\end{equation}
\begin{equation}
\small
    \overline{F}_t^l = \hat{F}_t^{l-1} + V*\hat{F}_t^{l} + b_4,
\end{equation}
where $\overline{F}_t^l$ is the output of the $t$-th segment in the $l$-th pyramid unit, $d$ denotes the dilated size, $\{W^i\}_{i=1}^3 \in \mathbb{R}^{D\times D}$ are convolution filter parameters, $b_3 \in \mathbb{R}^D$ is the bias vector, * denotes the $1\times1$ convolution operation, $V\in \mathbb{R}^{D\times D}$ and $b_4 \in \mathbb{R}^D$ are convolutional weights and bias.

We keep the dilation size of each unit equal to the interactive size of the fixed-size attention mechanism. Besides, the size of each unit is different, thereby guaranteeing that features obtained by these units contain contexts in multiple scales. Finally, we preserve the output of all units $\{F_v^i, F_a^i\}_{i=1}^L$ as the multimodal pyramid features, where $L$ is the total number of pyramid units.

\subsection{Adaptive Semantic Fusion Module} 
\label{subsec: adaptive}

A simple way to integrate pyramid features is to conduct pooling over the unit level. However, these pooling methods lack interactions between different levels, resulting in the incompatibility of multi-scale semantic information. To address this problem, we put forward an adaptive semantic fusion module, which consists of a unit-level attention block to explore the correlation of pyramid features and a selective fusion block for adaptively feature integration.

\noindent\textbf{Unit-level attention.} Since the interaction size of each pyramid unit is strictly restricted, pyramid units focus on areas in distinct scopes and generate semantic information at different levels, which results in a large semantic gap between pyramid features. To this end, we introduce the unit-level attention to provide contextual interactions and refine the pyramid features. The unit-level attention considers the similarity of contents in different units, and the interacting results vary with the characteristics of the captured features. The interacting process can be formulated as follows,
\begin{equation}
    r_t = att(\overline{F}_tW_{q}, \overline{F}_tW_{k}, \overline{F}_tW_{v}),
\end{equation}
where $\overline{F}_t\in \mathbb{R}^{L\times D}$ is the outputs of all pyramid units in the $t$-th segment, $W_{q}, W_{k}, W_{v}$ are learnable parameters.

\noindent\textbf{Selective fusion.} Since the type and length of events are uncertain, the model is supposed to pay more attention to features at suitable levels. Therefore, outputs of pyramid units are fused by a selective fusion block after building the relation-aware connections. Specifically, we perform a linear projection on each modality to gain the fusion weights of each unit. By doing so, the selective fusion block can dynamically assign weights on pyramid features in different granularities, and the fusion results vary with the characteristics and the event's type of the given video. The fused features are computed by the weighted summation,
\begin{equation}
    \hat{r}_t = \sum_{l=1}^L w^l_tr_t^l, 
\end{equation}
\begin{equation}
    w^l_t = \sigma(W_{sf}r_t^l+b_{sf}),
\end{equation}
where $W_{sf}$ and $b_{sf}$ denote the linear projection parameters, $\sigma$ denotes the sigmoid function. We use sigmoid instead of softmax since the characteristics of pyramid features are not mutually exclusive.

Finally, the probabilities in each modality can be computed by the sigmoid and the softmax function for multiple events and single event, respectively. For audio-visual events, since the event is both audible and visible in the same segment, the event probability in the $t$-th segment $p^t_{av}$ can be computed by the logical conjunction of uni-modal predictions, which is formulated as $p^t_{av} = p_a^t * p_v^t$. 

\section{Experiments}

\subsection{Audio-Visual Event Localization}

\begin{table}[t]
\centering
\caption{Overall accuracy (\%) compared with prior methods in both fully and weakly supervised manner. * denotes results re-implemented by the same feature extractor. Subset means the subset of the AVE dataset where events have multiple lengths and do not occur throughout the whole video.}
\begin{tabular}{@{}ccc@{}}
\toprule
Method     & Sup.(\%) & W-Sup.(\%) \\ \midrule
AVE~\cite{tian2018audio}        & 68.6                 & 66.7              \\
AVDSN~\cite{lin2019dual}        & 72.6                 & 67.3              \\
DAM~\cite{wu2019dual}           & 74.5                 & -                 \\
AVRB~\cite{ramaswamy2020see}    & 74.8                 & 68.9              \\
AVIN~\cite{ramaswamy2020makes}  & 75.2                 & 69.4              \\
CMAN~\cite{xuan2020cross}       & 73.3*                & 70.4*             \\
AVT~\cite{lin2020audiovisual}   & 76.8                 & 70.2              \\
MPN~\cite{yu2021mpn}            & 77.6                 & 72.0              \\
CMRAN~\cite{CMRAN2020Xu}        & 77.4                 & 72.9              \\
PSP~\cite{zhou2021positive}     & \textbf{77.8}        & \textbf{73.5}     \\ \midrule
MM-Pyramid (Ours)                      & \textbf{77.8}        & 73.2   \\ \midrule
MPN on Subset    & 62.1        & 53.8      \\
CMRAN on Subset  & 62.3        & 54.2      \\
PSP on Subset    & 62.7        & 54.4      \\ \midrule
MM-Pyramid (Ours) on Subset                     & \textbf{63.9} & \textbf{55.3}    \\ \bottomrule
\end{tabular}
\label{table1}
\end{table}

\noindent\textbf{Dataset and metrics.} Audio-Visual Event (AVE)~\cite{tian2018audio} dataset is an audio-visual event dataset with 4,183 video clips of 29 classes. Each video clip is 10 seconds with event category annotation per second. We follow the original setting as~\cite{tian2018audio} and divide the dataset as 80\%/10\%/10\% for training, validation, and test, respectively. The overall segment-wise accuracy is used for evaluation, which is the percentage of all matching segments. 

\noindent\textbf{Implementation details.} Since there is only one audio-visual event in a video, we decompose the task into video-level category predictions and segment-level relevance predictions as prior methods~\cite{wu2019dual, CMRAN2020Xu, yu2021mpn} do. Video-level categories are predicted by a temporal average pooling layer followed by a linear classifier, while segment-level relevance predictions are obtained by a segment-wise event-related binary classifier. We employ the VGG-19~\cite{simonyan2014very} network pre-trained on ImageNet~\cite{deng2009imagenet} and the VGGish~\cite{hershey2017cnn} network pre-trained on AudioSet~\cite{gemmeke2017audio} for feature extraction. We use Adam~\cite{kingma2014adam} as optimizer. The initial learning rate is 2e-5 and divided by 10 after 50 epochs. We complement more details in Appendix A.

\noindent\textbf{Comparison with the state-of-the-arts.} We compare our model with all prior methods as shown in~\cref{table1}. Results show that our model achieves comparable results with the state-of-the-art method PSP~\cite{zhou2021positive}. Since more than 60\% of events occur throughout the entire video in this dataset, the advantages of our model for detecting events of different lengths are not fully embodied in the full AVE dataset. Therefore, we conduct additional experiments on the subset where events are in multiple lengths. Results show that our model obtains higher performance on the multiple length events subset both in fully and weakly supervised settings. This proves our declaration that our model can detect more events with different lengths via the pyramid setting.

\subsection{Audio-Visual Video Parsing}

\begin{table}[t]
\caption{Audio-visual video parsing F-score results (\%) in comparison with recent weakly-supervised methods.}
\centering
\begin{tabular}{@{}c|c|cc@{}}
\toprule
Event Type                    & Methods          & \begin{tabular}[c]{@{}c@{}}Segment\\ Level\end{tabular} & \begin{tabular}[c]{@{}c@{}}Event\\ Level\end{tabular} \\ \midrule
\multirow{8}{*}{Audio}        & Kong et. al 2018~\cite{kong2018audio}   & 39.6                                & 29.1                   \\
                              & TALNet~\cite{wang2017spatiotemporal}    & 50.0                                & 41.7                   \\
                              & AVE~\cite{tian2018audio}                & 47.2                                & 40.4                   \\
                              & AVDSN~\cite{lin2019dual}                & 47.8                                & 34.1                   \\
                              & HAN~\cite{tian2020avvp}                 & 60.1                                & 51.3                   \\
                              & Ours                                    & \textbf{60.9}               & \textbf{52.7}  \\
                              & HAN+MA~\cite{wu2021exploring}           & 60.3	                              & 53.6                   \\
                              & Ours+MA                                 & \textbf{61.1}               & \textbf{53.8}  \\ \midrule
\multirow{8}{*}{Visual}       & STPN~\cite{nguyen2018weakly}            & 46.5                                & 41.5                   \\
                              & CMCS~\cite{liu2019completeness}         & 48.1                                & 45.1                   \\
                              & AVE~\cite{tian2018audio}                & 37.1                                & 34.7                   \\
                              & AVDSN~\cite{lin2019dual}                & 52.0                                & 46.3                   \\
                              & HAN~\cite{tian2020avvp}                 & 52.9                                & 48.9                    \\ 
                              & Ours                                    & \textbf{54.4}               & \textbf{51.8}   \\
                              & HAN+MA~\cite{wu2021exploring}           & 60.0	                              & 56.4                    \\
                              & Ours+MA                                 & \textbf{60.3}               & \textbf{56.7}   \\ \midrule
\multirow{6}{*}{Audio-Visual} & AVE~\cite{tian2018audio}                & 35.4                                & 31.6                    \\
                              & AVDSN~\cite{lin2019dual}                & 37.1                                & 26.5                     \\
                              & HAN~\cite{tian2020avvp}                 & 48.9                                & 43.0                    \\
                              & Ours                                    & \textbf{50.0}               & \textbf{44.4}     \\
                              & HAN+MA~\cite{wu2021exploring}           & 55.1	                              & 49.0                      \\
                              & Ours+MA                                 & \textbf{55.8}               & \textbf{49.4}     \\ \midrule
\multirow{6}{*}{Type@AV}      & AVE~\cite{tian2018audio}                & 39.9                                & 35.5                      \\
                              & AVDSN~\cite{lin2019dual}                & 45.7                                & 35.6                      \\
                              & HAN~\cite{tian2020avvp}                 & 54.0                                & 47.7                       \\
                              & Ours                                    & \textbf{55.1}               & \textbf{49.9}      \\ 
                              & HAN+MA~\cite{wu2021exploring}           & 58.9	                              & 53.0                       \\
                              & Ours+MA                                 & \textbf{59.7}               & \textbf{54.1}      \\ \midrule
\multirow{6}{*}{Event@AV}     & AVE~\cite{tian2018audio}                & 41.6                                & 36.5                       \\
                              & AVDSN~\cite{lin2019dual}                & 50.8                                & 37.7                       \\
                              & HAN~\cite{tian2020avvp}                 & 55.4                                & 48.0                       \\
                              & Ours                                    & \textbf{57.6}               & \textbf{50.5}      \\ 
                              & HAN+MA~\cite{wu2021exploring}           & 57.9	                              & 50.6                       \\
                              & Ours+MA                                 & \textbf{59.1}               & \textbf{51.2}      \\ \bottomrule
\end{tabular}
\label{table2}
\end{table}

\noindent\textbf{Dataset and metrics.} Look, Listen, and Parse (LLP)~\cite{tian2020avvp} dataset derived from AudioSet~\cite{gemmeke2017audio} is constructed for the audio-visual video parsing task. It contains 11,849 videos of 25 event categories. Each video clip is 10s long and contains 1.64 events on average. 1,849 videos are randomly sampled to be annotated at the event level for evaluation, where the validation and test set includes 649 and 1,200 videos, respectively. The remaining 10,000 videos are annotated in video-level for training. Following~\cite{tian2020avvp}, we employ the segment-level and event-level F-scores of all modalities as the evaluation metrics. The segment-level metric computes the F-score of each segment, and the event-level metric computes the event-level F-score by comparing the concatenated positive consecutive segments with the event-level ground-truth, where the mIOU is set as 0.5. Furthermore, two average metrics Type@AV and Event@AV are also reported. Type@AV means computing the F-score of each event type (audio, visual, and audio-visual) and averaging these results. Event@AV is generated by considering all types of events in each video and computing the composite F-score results.

\begin{table*}[t]
\caption{Ablation studies with different components on the audio-visual video parsing task. We propose several variants to investigate the impact of multimodal pyramid setting, attentive feature pyramid module, and adaptive semantic fusion module.}
\centering
\begin{tabular}{@{}c|cc|cc|cc|cc|cc@{}}
\toprule
\multirow{2}{*}{Model}     & \multicolumn{2}{c|}{Audio}    & \multicolumn{2}{c|}{Visual}   & \multicolumn{2}{c|}{Audio-Visual} & \multicolumn{2}{c|}{Type@AV}  & \multicolumn{2}{c}{Event@AV} \\ \cmidrule(l){2-11} 
                           & Seg       & Eve         & Seg       & Eve         & Seg         & Eve           & Seg       & Eve         & Seg       & Eve         \\ \midrule
MM-Pyramid-Last        & 59.6          & 51.4          & 53.6         & 50.1          & 49.2           & 43.2            & 54.1          & 48.2          & 56.4          & 48.6          \\ \midrule
MM-Unpyramid & 59.7          & 49.6          & 53.3          & 50.0          & 47.7            & 41.5            & 53.6          & 47.0          & 57.4          & 48.1          \\ \midrule
Hybr-Trans w/ PE & 60.1          & 51.9          & 53.0          & 50.1          & 48.4            & 43.7            & 54.5          & 47.6          & 56.1          & 48.5          \\ \midrule
MM-Pyramid w/o conv     & 60.0          & 50.8          & 52.4          & 49.2          & 47.9            & 41.7            & 53.1          & 47.0          & 56.4          & 47.8          \\ \midrule
MM-Pyramid w/o residual      & 60.4          & 51.5          & 52.5          & 49.5          & 47.7            & 41.8            & 53.5          & 47.6          & 57.1          & 48.7          \\ \midrule
MM-Pyramid w/o ULA & 60.6          & 52.1          & 53.4          & 49.8          & 48.8            & 43.6            & 54.3          & 48.5          & 56.8          & 49.2          \\ \midrule
MM-Pyramid w/o SF & 60.5          & 51.8          & 53.6          & 49.9          & 48.7            & 43.1            & 54.3          & 48.3          & 56.9          & 49.1          \\ \midrule
MM-Pyramid (full)              & \textbf{60.9} & \textbf{52.7} & \textbf{54.4} & \textbf{51.8} & \textbf{50.0}   & \textbf{44.4}   & \textbf{55.1} & \textbf{49.9} & \textbf{57.6} & \textbf{50.5} \\ \bottomrule
\end{tabular}
\label{table3}
\end{table*}

\noindent\textbf{Implementation details.} The outputs of MM-Pyramid $p^t_v$, $p^t_a$, and $p^t_{av}$ represent the uni-modal and multimodal parsing results. Since this task is performed in a weakly supervised manner, we follow \cite{tian2020avvp} to leverage an attentive MMIL pooling to generate video-level predictions. We also use label smoothing~\cite{szegedy2016rethinking} to alleviate label noises of the weakly supervised setting. Following~\cite{tian2020avvp}. we employ ResNet-152~\cite{he2016deep} and R(2+1)D~\cite{tran2018closer} to extract visual features, and VGGish network to extract audio features. We use Adam~\cite{kingma2014adam} optimizer and set the learning rate as 1e-4, which is degraded by a factor of 5 after 10 epochs. More details are listed in Appendix B.

\noindent\textbf{Comparison with the state-of-the-arts.} We compare with the state-of-the-art methods HAN~\cite{tian2020avvp}, AVE~\cite{tian2018audio} and AVDSN~\cite{lin2019dual}, as well as several competitive weakly supervised event detection methods, which is shown in~\cref{table2}. To be specific, we choose temporal action localization methods STPN~\cite{nguyen2018weakly} and CMCS~\cite{liu2019completeness}, sound event detection methods TALNet~\cite{wang2019comparison} and Kong et al.~\cite{kong2018audio}. For the new modality-aware method MA~\cite{wu2021exploring}, since they do not conduct optimization from the network perspective and use the same hybrid attention network (HAN) as~\cite{tian2020avvp}, our method is not mutually exclusive with their strategy. Therefore, we provide results both using the raw training strategy and the new label refinement and contrastive learning strategy. Results show that our model outperforms baseline methods on all evaluation metrics in a large margin. For the raw training strategy, our model yields up to 2.9\% higher on the unimodal metrics (Visual\&Event-level) and up to 2.5\% higher on the multimodal metrics (Event@AV\&Event-level). This proves that the insight of capturing and integrating multimodal pyramid features enables the localization of events in multiple lengths precisely, which further results in better video parsing performance.

\subsection{Ablation Studies}

\noindent\textbf{Do pyramid units help?} We first investigate the impact of the pyramid units. As shown in~\cref{table3}, ``MM-Pyramid-Last" means only the output of the last pyramid unit is preserved. ``MM-Unpyramid" denotes the sizes of all pyramid units are identical and equal to the size of the last pyramid units in the raw framework. ``Hybr-Trans w/PE`` denotes the 4-layer hybrid transformer encoder with positional encoding. Results show that our full model outperforms all ablated models, indicating the significance of capturing multi-level features via our pyramid settings. We argue that the multimodal information learned by different levels of contexts resolves videos at multiple granularities. We also make comparisons among different transformer-based structures, which are shown in Appendix B.

\noindent\textbf{Does dilated residual convolution help?} We also explore the efficacy of our dilated convolution block. ``MM-Pyramid w/o conv" indicates that the entire dilated block is removed, ``MM-Pyramid w/o residual" means the dilated residual block is replaced with a single $3\times3$ convolution layer. Without dilated convolution block, the performance declines significantly, showing the integration of pyramid features is necessary for semantics information. The performance also declines when using the vanilla convolution layer, proving that the residual connections and $1\times1$ convolution can enhance the expressiveness of integrated features.

\noindent\textbf{Does adaptive semantic fusion module help?} Multimodal pyramid features are fused interactively. To reveal the contribution of our adaptive semantic fusion module, we propose two ablated models ``MM-Pyramid w/o ULA" and ``MM-Pyramid w/o SF". The first model is constructed by removing the unit-level attention, while the other fuses pyramid features via an average pooling layer instead of the selective fusion block. The performance of ``MM-Pyramid w/o ULA" declines, which indicates that building the relation-aware connections of multi-scale features is effective. Our model also outperforms ``MM-Pyramid w/o SF". We argue that this proves the insight of integrating pyramid features selectively helps the acquisition of complete video scene understanding.

\noindent\textbf{Impact of parameter sharing setting} To investigate the performance of the parameter sharing strategy of cross-modal attention blocks, we complement a non-sharing ablated model as shown in~\cref{table4}. The result indicates that the sharing matrices get comparable performance with lower computing complexity compared with the non-sharing setting.

\begin{table}[htbp]
\centering
\caption{Segment-level and event-level f1-scores(\%) comparison with different parameter sharing strategy.}
\begin{tabular}{@{}ccccc@{}}
\toprule
\multirow{2}{*}{Model} & \multicolumn{2}{c}{Type@AV} & \multicolumn{2}{c}{Event@AV} \\ \cmidrule(l){2-5} 
                                     & Segment       & Event       & Segment        & Event       \\ \midrule
Non-Sharing                          & 54.8          & \textbf{50.1}        & 57.2           & 50.0         \\ 
Sharing (\textbf{ours})              & \textbf{55.1}          & 49.9        & \textbf{57.6}           & \textbf{50.5}        \\ \bottomrule
\end{tabular}
\label{table4}
\end{table}

\subsection{Qualitative Results}

\begin{figure}[htbp]
\includegraphics[width=0.47\textwidth]{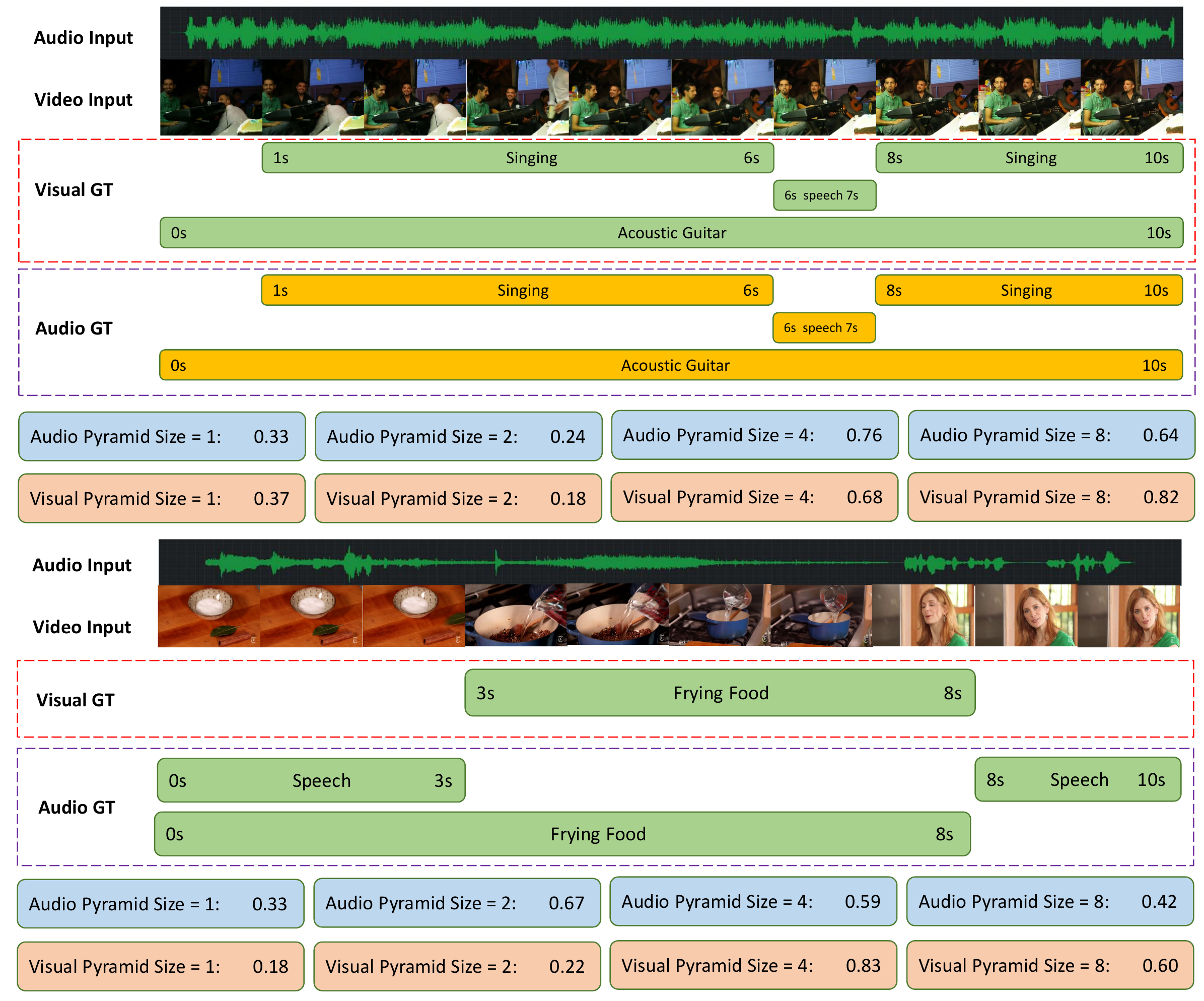}
\caption{Qualitative results of the selective fusion, which assigns each pyramid unit weights for feature integration.}
\label{figure5}
\end{figure}

\begin{figure*}[h]
\centering
\includegraphics[width=0.95\textwidth]{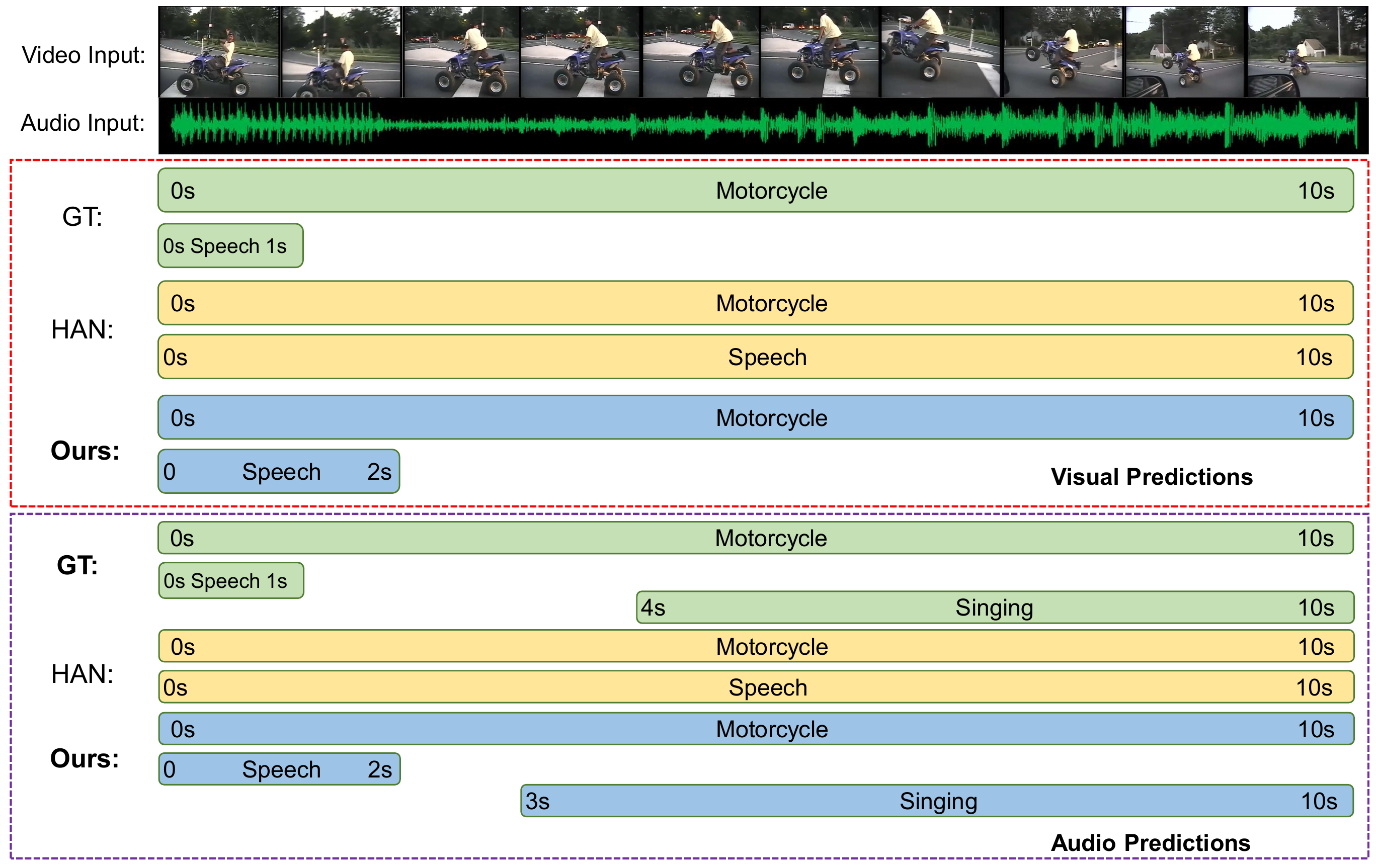}
\caption{Qualitative comparison with the weakly-supervised audio-visual video parsing method HAN. The red dotted box includes the visual predictions, and the audio predictions are in the purple dotted box. The green, yellow, and blue labels denote the ground-truth, predictions of HAN, and predictions of our MM-Pyramid, respectively.}
\label{figure6}
\end{figure*}

\noindent\textbf{Role of selective fusion.} To illustrate the effectiveness of our selective fusion block in the attentive semantic fusion module, we conduct qualitative results as shown in~\cref{figure6}. The sample video in the picture consists of some long events as well as a small audio-visual event ``speech''. It should be noticed that since the characteristics of pyramid features are not mutually exclusive, we use the sigmoid function as the substitution of softmax to generate fusion weights, thus the sum of weights is not 1. Results show that the selective fusion module assigns relatively high scores on the pyramid units of large scales, which indicates the effectiveness of our feature integration method. The sample video below the picture consists of one visual event of medium length yet several audio events in miscellaneous lengths. Therefore, the selective fusion block focuses more on the visual pyramid units with medium lengths and disperses weights into all audio pyramid units.

\noindent\textbf{Capability of detecting events in multiple lengths.} We also illustrate our model's capability of capturing multiple events. As shown in~\cref{figure5}, we conduct qualitative experiments on the audio-visual video parsing task in comparison with HAN~\cite{tian2020avvp}. The green labels are the ground truths, and the yellow and blue labels indicate the predictions of HAN and our model, respectively. We found that though the HAN model can precisely predict events that exist throughout the whole video, it fails to detect a short-term event (singing audio events from 4th to 10th seconds) and predicts incorrect event temporal boundary (the speech event in the first second). However, our MM-Pyramid model tends to recognize all events of different sizes and provide predictions with only small deviations (one-second errors in speech and singing events). This result reveals that our model is capable of exploring features in different granularities, which further leads to localizing events in diverse lengths precisely. We provide more qualitative results in Appendix E, including additional visualization results and error analysis.

\section{Limitation}
Though our MM-Pyramid framework shows the efficacy of detecting multiple events in different lengths, the advantage is limited when detecting events that occur throughout the whole video compared with other methods. This can be shown in the experimental results of the audio-visual event localization task, in which task the majority (66.4\%) of events span over the whole video. In that situation, we suggest injecting our model or our multimodal pyramid feature methodology into other single-shot event detection methods as an enhancement for detecting multiple events. To this end, finding a flexible way to assemble our proposed multimodal pyramid paradigm with some widely-adopted temporal localization methods could be a promising research direction.

\section{Conclusion}

In this paper, we propose a novel Multimodal Pyramid Attentional Network (MM-Pyramid) for audio-visual event localization and weakly-supervised audio-visual video parsing. Our model captures and integrates multimodal pyramid features in distinct temporal scales for comprehensive scene understanding. To acquire features in different granularities, we propose a novel attentive feature pyramid module, which is composed of the fixed-size attention mechanism and dilated convolution block. Furthermore, we propose an adaptive semantic fusion module to refine and fuse pyramid features in an interactive and selective way. Extensive experiments on the AVE and LLP datasets demonstrate the effectiveness of our proposed approach on localizing events in multiple lengths. In future works, we plan to expand our multimodal pyramid architecture to more audio-visual scenarios such as violence detection, representation learning, and multimodal reasoning. 

\begin{acks}
This work was supported by National Natural Science Foundation of China (No. 62172101, No. 61976057). This work was supported (in part) by the Science and Technology Commission of Shanghai Municipality (No. 21511101000, No. 21511100602), and SPMI Innovation and Technology Fund Projects (SAST2020-110)
\end{acks}

\clearpage
\bibliographystyle{ACM-Reference-Format}
\bibliography{reference}

\end{document}